\documentclass[conference]{IEEEtran}
\IEEEoverridecommandlockouts
\usepackage{cite}
\usepackage{amsmath,amssymb,amsfonts}
\usepackage{algorithmic}
\usepackage{graphicx}
\usepackage{textcomp}
\def\BibTeX{{\rm B\kern-.05em{\sc i\kern-.025em b}\kern-.08em
    T\kern-.1667em\lower.7ex\hbox{E}\kern-.125emX}}

\usepackage{tabularx}
\usepackage{threeparttable}
\newcommand*{\Scale}[2][4]{\scalebox{#1}{$#2$}}%
\usepackage[]{dashrule, colortbl}
\usepackage[utf8]{inputenc} % allow utf-8 input
\usepackage[T1]{fontenc}    % use 8-bit T1 fonts
\usepackage{hyperref}       % hyperlinks
\usepackage{url}            % simple URL typesetting
\usepackage{booktabs}       % professional-quality tables
\usepackage{amsfonts}       % blackboard math symbols
\usepackage{nicefrac}       % compact symbols for 1/2, etc.
\usepackage{microtype}      % microtypography
\usepackage{amsmath,xparse, graphicx, soul}
\mathchardef\mhyphen="2D % Define a "math hyphen"
\usepackage{subfig}
\usepackage{xfakebold}
\usepackage{float}

\usepackage{booktabs, makecell} % For formal tables
\usepackage{pifont}
\definecolor{blue}{rgb}{0.2, 0.4, 1.0}
\definecolor{orange}{rgb}{0.6, 0.4, 0.2}
\newcommand{\cmark}{\textcolor{blue}{\ding{51}}}
\newcommand{\xmark}{\textcolor{orange}{\ding{55}}}
\usepackage{indentfirst}

\usepackage[linesnumbered,boxed,ruled,commentsnumbered]{algorithm2e}
\usepackage{times}  %Required
\usepackage{helvet}  %Required
\usepackage{courier}  %Required
\usepackage{url}  %Required
\usepackage{graphicx}  %Required

\usepackage{multirow}
\usepackage{amsmath,bm}
\usepackage{enumerate}
\usepackage[dvipsnames]{xcolor}
\usepackage{caption}
\usepackage{etoolbox}

\usepackage[colorinlistoftodos]{todonotes}

\usepackage{xspace}
\newcommand{\Transformer}{\textsc{transformer}\xspace}
\newcommand{\TransformerSS}{\textsc{\scriptsize transformer}\xspace}
\newcommand{\Transformernospace}{\textsc{transformer}}
\newcommand{\TransformernospaceSS}{\textsc{\scriptsize transformer}}

\newcommand{\metavecSS}{\textsc{\scriptsize metapath2vec}\xspace}

\newcommand{\HERecSS}{\textsc{\scriptsize HERec}\xspace}

\newcommand{\HINVecSS}{\textsc{\scriptsize hin2vec}\xspace}

\newcommand{\HeGANSS}{\textsc{\scriptsize HeGAN}\xspace}

\newcommand{\HetGNN}{\textsc{\small HetGNN}\xspace}
\newcommand{\HetGNNSS}{\textsc{\scriptsize HetGNN}\xspace}

\newcommand{\RGCN}{\textsc{\small R-GCN}\xspace}
\newcommand{\RGCNSS}{\textsc{\scriptsize R-GCN}\xspace}
\newcommand{\RGCNnospace}{\textsc{\small R-GCN}}
\newcommand{\RGCNnospaceSS}{\textsc{\scriptsize R-GCN}}

\newcommand{\BAGNN}{\textsc{\small BA-GNN}\xspace}
\newcommand{\BAGNNSS}{\textsc{\scriptsize BA-GNN}\xspace}
\newcommand{\BAGNNnospace}{\textsc{\small BA-GNN}}
\newcommand{\BAGNNnospaceSS}{\textsc{\scriptsize BA-GNN}}

\newcommand{\BAGNNnode}{\textsc{\small BA-GNN-node}\xspace}
\newcommand{\BAGNNnodeSS}{\textsc{\scriptsize BA-GNN-node}\xspace}
\newcommand{\BAGNNrel}{\textsc{\small BA-GNN-relation}\xspace}
\newcommand{\BAGNNrelSS}{\textsc{\scriptsize BA-GNN-relation}\xspace}

\newcommand{\BAGNNhan}{\textsc{\small BA-GNN(node)+HAN(rel.)}\xspace}
\newcommand{\BAGNNhanSS}{\textsc{\scriptsize BA-GNN(node)+HAN(rel.)}\xspace}

\newcommand{\HAN}{\textsc{\small HAN}\xspace}
\newcommand{\HANnospace}{\textsc{\small HAN}}
\newcommand{\HANSS}{\textsc{\scriptsize HAN}\xspace}

\newcommand{\GCN}{\textsc{\small GCN}\xspace}
\newcommand{\GCNSS}{\textsc{\scriptsize GCN}\xspace}
\newcommand{\GCNnospace}{\textsc{\small GCN}}

\newcommand{\GAT}{\textsc{\small GAT}\xspace}
\newcommand{\GATSS}{\textsc{\scriptsize GAT}\xspace}

\newcommand{\TransE}{\textsc{\small TransE}\xspace}
\newcommand{\TransESS}{\textsc{\scriptsize TransE}\xspace}

\newcommand{\HolE}{\textsc{\small HolE}\xspace}
\newcommand{\HolESS}{\textsc{\scriptsize HolE}\xspace}

\newcommand{\DistMult}{\textsc{\small DistMult}\xspace}
\newcommand{\DistMultSS}{\textsc{\scriptsize DistMult}\xspace}

\newcommand{\ComplEx}{\textsc{\small ComplEx}\xspace}
\newcommand{\ComplExSS}{\textsc{\scriptsize ComplEx}\xspace}

\newcommand{\GTNSS}{\textsc{\scriptsize GTN}\xspace}

\newcommand{\DySAT}{\textsc{\small DySAT}\xspace}
\newcommand{\DySATSS}{\textsc{\scriptsize DySAT}\xspace}

\newcommand{\TGAT}{\textsc{\small TGAT}\xspace}
\newcommand{\TGATSS}{\textsc{\scriptsize TGAT}\xspace}

\newcommand{\HGT}{\textsc{\small HGT}\xspace}
\newcommand{\HGTSS}{\textsc{\scriptsize HGT}\xspace}

\newcommand{\TemporalGAT}{\textsc{\small TemporalGAT}\xspace}
\newcommand{\TemporalGATSS}{\textsc{\scriptsize TemporalGAT}\xspace}

\begin{document}

\title{Bi-Level Attention Graph Neural Networks}

% \begin{tabular}[t]{c}\bf\rule{\z@}{24\p@}
%             Roshni G. Iyer\quad Yizhou Sun\quad Wei Wang\\
%             \small{University of California, Los Angeles}\\
%             \small{Los Angeles, CA, 90095, USA}\\
%             \small{\{roshnigiyer, yzsun, weiwang\}@cs.ucla.edu}\\
%         \end{tabular}%
%         \begin{tabular}[t]{c}\bf\rule{\z@}{24\p@}
%             Justin Gottschlich\\
%             \small{Intel Labs \& University of Pennsylvania, USA}\\
%             \small{Santa Clara, CA, 95054, USA}\\
%             \small{justin.gottschlich@intel.com}\\
%         \end{tabular}
% \author{Roshni G. Iyer}

\author{Roshni G. Iyer \\
\IEEEauthorblockA{\textit{University of California, Los Angeles} \\
% Los Angeles, CA, 90095, USA  \\
roshnigiyer@cs.ucla.edu}
\and
\IEEEauthorblockN{Wei Wang}
\IEEEauthorblockA{\textit{University of California, Los Angeles} \\
% Los Angeles, CA, 90095, USA \\
weiwang@cs.ucla.edu}
\and
\IEEEauthorblockN{Yizhou Sun}
\IEEEauthorblockA{\textit{University of California, Los Angeles} \\
yzsun@cs.ucla.edu}
% \and
% \and
% \and
}

\maketitle

\begin{abstract}
Recent graph neural networks (GNNs) with the attention mechanism have historically been limited to small-scale homogeneous graphs (HoGs). However, GNNs handling heterogeneous graphs (HeGs), which contain several entity and relation types, all have shortcomings in handling attention. 
Most GNNs that learn graph attention for HeGs learn either node-level or relation-level attention, but not both, limiting their ability to predict both important entities and relations in the HeG. Even the best existing method that learns both levels of attention has the limitation of assuming graph relations are independent and that its learned attention disregards this dependency association. To effectively model both multi-relational and multi-entity large-scale HeGs, we present Bi-Level Attention Graph Neural Networks (\BAGNN), scalable neural networks (NNs) that use a novel bi-level graph attention mechanism. \BAGNN models both node-node and relation-relation interactions in a personalized way, by hierarchically attending to both types of information from local neighborhood contexts instead of the global graph context. Rigorous experiments on seven real-world HeGs show \BAGNN consistently outperforms all baselines, and demonstrate quality and transferability of its learned relation-level attention to improve performance of other GNNs.
\end{abstract}


\setlength{\parindent}{3ex}

\vspace{-6,5mm}
\section{Introduction}
\label{introduction}
\begin{figure}[h]
    \centering
    \includegraphics[width=\linewidth]{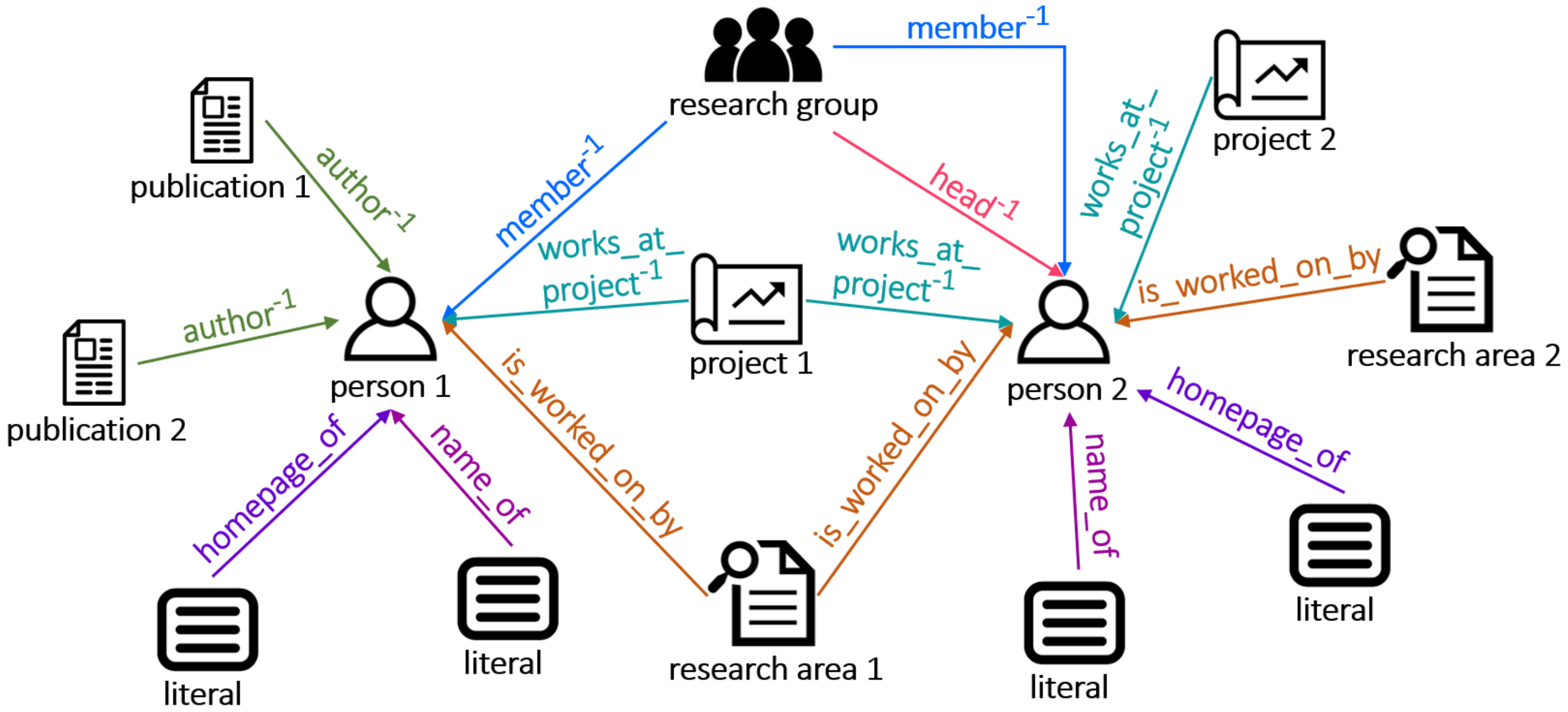}
    \caption{\small{Partial HeG of AIFB dataset.}}
    \label{fig:aifb-example}
    % \vspace{-7mm}
\end{figure}

Highly multi-relational data are characteristic of real-world HeGs. Relational data in HeGs are defined as triples of form \textsl{(h:head entity, r:relation, t:tail entity)}, indicating that two entities are connected by a specific relation type. Figure~\ref{fig:aifb-example} shows a HeG formed by such triples.
% e.g., \textsl{(research area 1, is\_worked\_on\_by, person 1)}.
However, even comprehensive HeGs~\cite{freebase}
% such as Freebase~\cite{freebase}, DBpedia~\cite{db}, and YAGO~\cite{yago} 
remain incomplete.
% and difficult to maintain. 
Regarding HeGs completion, despite the recent years' research progress in developing GNNs for representation learning in various domains~\cite{HeGAN, HGT, HERec} and adapting the successful attention mechanism~\cite{Transformer,GAT},
% Research has newly focused on developing GNNs for representation learning in various domains, e.g. social networks~\cite{HeGAN, HGT, HERec}. Recent works have advanced GNNs through fundamental operations including NN convolutions~\cite{GCN} and graph attention~\cite{GAT}. 
% Due to the large success of attention in the natural language processing (NLP) domain~\cite{Transformer}, works have extended this mechanism to the graph domain~\cite{GAT}. Despite their achievements, 
most GNNs face several challenges. They either are ill-equipped to handle HeGs~\cite{GCN, GAT}, or do handle HeGs but do not learn graph attention~\cite{metapath2vec, rgcn, HIN2Vec, HetGNN}, or learn inaccurate graph attention~\cite{TemporalGAT, DySAT, HAN, GTN}.

Considering the GNNs that learn graph attention, their architectures are limited to only one level of attention, either for nodes or relations, but rarely for both, shown in Table~\ref{table:model-limitations}. This is problematic for modeling HeGs which contain several different entity and relation types. Bi-level attention is more powerful in learning compared to uni-level attention, where only one level of attention is learned by the model. Bi-level attention learns attention at different levels of granularity in HeGs which captures more information about graph components than a uni-level attention mechanism is capable of. \HAN, one of the few models that attempts to use bi-level attention, unsurprisingly falls short of capturing the associations between the node and relation levels in the HeG. First, \HAN places unnatural assumptions on the data because it treats graph relations as independent from each other, omitting most relation-relation interactions in HeGs. Second, it requires manually chosen meta paths that force many node-node and node-relation interactions to also be left out, and requires domain specific knowledge to compute. Third, \HAN lacks a general framework for systematically studying bi-level attention.     

To address the above challenges, in this paper, we present \textbf{B}i-Level \textbf{A}ttention \textbf{G}raph \textbf{N}eural \textbf{N}etworks (\BAGNN) for HeGs. To summarize, our work makes the following contributions: 

\begin{enumerate}
\renewcommand\labelenumi{(\theenumi)}
\vspace{-1mm}
\item We design a general framework for bi-level attention, and identify challenges of state-of-art NNs for HeGs.\label{item:1}

\vspace{-0.5mm}
\item We propose \BAGNN to model both multi-relational and multi-entity large-scale HeGs. \BAGNN avoids manually chosen meta paths, and learns personalized graph properties by integrating graph entity/relation types, graph structure, and graph attention using local graph neighborhoods instead of global graph context. \BAGNN improves accuracy of state-of-art GNNs and scales to million-node/edge graphs, like the AM archaeological dataset. \label{item:2}

\vspace{-4mm}
\item To our knowledge, we are the first to propose efficient bi-level attention GNNs that learn from dependency interactions of both nodes/relations and without meta paths. 
\label{item:3}  

\vspace{-5mm}
\item We rigorously experiment on seven real-world HeGs showing \BAGNN consistently outperforms major state-of-art NN groups, and also demonstrate quality and transferability of \BAGNNnospace\textnormal{’s} attention-induced change in graph structure to enrich other GNNs. \label{item:4}
\end{enumerate} 

\vspace{-1mm}
The remainder of this paper is organized as follows. Section~\ref{related-work} examines preliminaries and related work. Section~\ref{BR-GCN-architecture} presents a general framework for computing bi-level attention, and describes \BAGNNnospace\textnormal{'s} architecture. Section~\ref{evaluation} presents experiment results, ablation studies, and case studies of \BAGNN models, and section~\ref{conclusions} concludes.

\vspace{-1.5mm}
\section{Preliminary and Related Work}
\vspace{-1mm}
Here, we introduce HeG concepts and discuss the achievement of various state-of-art NNs, summarized in Table~\ref{table:model-limitations}.

% \vspace{-3mm}
\paragraph*{\textbf{\textit{{\normalsize Definition 1}}}} \textbf{Heterogeneous Graph:} We define HeGs as $\mathcal{G} = (\mathcal{V}, \mathcal{E})$ with nodes $v_{i} \in \mathcal{V}$, and edges $e_{i,j} \in \mathcal{E}$ connecting source $v_{i}$ and target $v_{j}$. Nodes in the HeG are associated with entity types through an entity mapping function $\Lambda(v_{i}) : \mathcal{V} \xrightarrow{} \mathcal{B}, \mathcal{B} = \{b | b \in \mathcal{B}\}$, for entity type $b$. Edges in the HeG are associated with relation types through a relation mapping function $\Gamma(e_{i,j}) : \mathcal{E} \xrightarrow{} \mathcal{R}, \mathcal{R} = \{r | r \in \mathcal{R}\}$ for relation type $r$. For efficiency in our model, we compute entity and relation mapping functions for node and relation neighborhoods rather than globally such that $\Lambda(v_{i}) : \mathcal{V}_{i} \xrightarrow{} \mathcal{B}_{i}, \mathcal{B}_{i} \subset \mathcal{B}$ and $\Gamma(e_{i,j}) : \mathcal{E}_{i} \xrightarrow{} \mathcal{R}_{i}, \mathcal{R}_{i} \subset \mathcal{R}$. In this paper, we consider local HeG neighborhoods of a node $v_{i} \in \mathcal{V}$ consisting of both one-hop nodes, $\{v_{j} | v_{j} \in \mathcal{V}_{i}\}$ and one-hop relations, $\{r|r \in \mathcal{R}_{i}\}$.

% \vspace{-3mm}
\paragraph*{\textbf{\textit{{\normalsize Definition 2}}}} \textbf{Meta Relation:} The meta relation for $e_{i,j}$ between source $v_{i}$ and target $v_{j}$ is $(\Lambda(v_{i}), \Gamma(e_{i,j}), \Lambda(v_{j}))$, and $\Gamma(e_{i,j})^{-1} = \Gamma(e_{j,i})$ is the inverse of $\Gamma(e_{i,j})$. In this paper, we loosely use the term \textit{relation} to denote meta relation. Traditional meta paths are a sequence of such meta relations. 

% \vspace{-3mm}
\paragraph*{\textbf{\textit{{\normalsize Definition 3}}}} \textbf{Graph Attention:} Graph attention enables NNs to learn useful graph representations by selectively attending to different nodes and relations. Multiplicative and additive attention are state-of-art attention mechanisms used in NNs~\cite{Transformer, GAT}, both of which operate on encoder states. Multiplicative attention uses an inner product or cosine similarity of encoder states while additive attention is a linear combination or concatenation of encoder states.

\label{related-work}

 \begin{table}[t!]
 \scriptsize
  \caption{\textmd{\small{Properties of GNN and attention-based models, with \cmark\ as advantages, and \xmark\ as disadvantages.}}}
  \label{table:model-limitations}
  \newcolumntype{L}{>{\centering\arraybackslash}X}
  \centering
  \begin{tabularx}{\linewidth}{|p{0.4cm}|c@{\hskip 0.5mm}|p{0.35cm}|p{0.35cm}|p{0.35cm}|p{0.35cm}|p{0.35cm}|p{0.35cm}|p{0.35cm}|}
  \cline{1-9} \\[-2.4ex]
  Type & Model & [A] & 
 [B] & 
 [C] & 
 [D] &
 $^{\dagger}$[E] &
 [F] &
 [G]\\ [0.2ex]
\cline{1-9} \\[-2.5ex]
 -- & \TransformerSS~\cite{Transformer} & \cmark & \cmark & \cmark & \xmark & \cmark & \cmark & --\\ [0.2 ex]
 \cline{1-9} \\[-2.5ex]
 & \TransESS~\cite{TransE} & \cmark & \cmark & \xmark & \xmark & \xmark & \cmark & \xmark\\[0.2ex]
 \cline{2-9} \\[-2.5ex]
 (1) %Non-GNN-based
 & \HolESS~\cite{HolE} & \cmark & \cmark & \xmark & \xmark & \xmark & \cmark & \xmark\\[0.2ex]
 \cline{2-9} \\[-2.5ex]
 %KGE models for HeGs
 & \DistMultSS~\cite{Distmult} & \cmark & \cmark & \xmark & \xmark & \xmark & \cmark & \xmark\\[0.2ex]
  \cline{2-9} \\[-2.5ex]
  & \ComplExSS~\cite{ComplEx} & \cmark & \cmark & \xmark & \xmark & \xmark & \cmark & \xmark\\[0.2ex]
 \cline{1-9} \\[-2.5ex]
 (2) %GNNs for
 & \GCNSS~\cite{GCN}& \xmark & \cmark & \xmark & \xmark & \xmark & \cmark & \xmark\\[0.2ex]
 \cline{2-9} \\[-2.5ex]
  %HoGs 
  & $\textmd{\GATSS}^{*}$~\cite{GAT} & \xmark & \cmark & \cmark & \xmark & \cmark & \cmark & \xmark\\[0.2ex]
 \cline{1-9} \\[-2.5ex]
 & \metavecSS~\cite{metapath2vec} & \cmark & \cmark & \xmark & \xmark & \xmark & \xmark & \xmark\\ [0.2 ex]
 \cline{2-9} \\[-2.5ex]
  & \HERecSS~\cite{HERec} & \cmark & \cmark & \xmark & \xmark & \xmark & \xmark & \xmark\\ [0.2 ex]
 \cline{2-9} \\[-2.5ex]
  & \HINVecSS~\cite{HIN2Vec} & \cmark & \cmark & \xmark & \xmark & \xmark & \xmark & \xmark\\ [0.2 ex]
 \cline{2-9} \\[-2.5ex]
  (3A) %Non-\TransformernospaceSS\textnormal{-} 
  & \HeGANSS~\cite{HeGAN} & \cmark & \cmark & \xmark & \xmark & \xmark & \cmark & \xmark\\ [0.2 ex]
 \cline{2-9} \\[-2.5ex]
%  based GNNs for HeGs 
 & \TemporalGATSS~\cite{TemporalGAT} & \xmark & \xmark & \cmark & \xmark & \cmark & \cmark & \xmark\\[0.2ex]
 \cline{2-9} \\[-2.5ex]
  & \HetGNNSS~\cite{HetGNN} & \cmark & \cmark & \xmark & \xmark & \xmark & \cmark & \xmark\\ [0.2 ex]
 \cline{2-9} \\[-2.5ex]
  & $\textmd{\RGCNSS}^{*}$~\cite{rgcn} & \cmark & \cmark & \xmark & \xmark & \xmark & \cmark & \xmark\\ [0.2 ex]
 \cline{2-9} \\[-2.5ex]
  & $\textmd{\HANSS}^{*}$~\cite{HAN} & \cmark & \cmark & \cmark & \cmark & \xmark & \xmark & \xmark\\[0.2ex]
 \cline{1-9} \\[-2.5ex]
 & \DySATSS~\cite{DySAT} & \cmark & \xmark & \cmark & \xmark & \cmark & \cmark & \cmark\\[0.2ex]
 \cline{2-9} \\[-2.5ex]
 (3B) %\TransformernospaceSS\textnormal{-}
 & \TGATSS~\cite{TGAT} & \xmark & \xmark & \cmark & \xmark & \cmark & \cmark & \cmark\\[0.2ex]
 \cline{2-9} \\[-2.5ex]
%  based GNNs for HeGs 
 & \HGTSS~\cite{HGT} & \cmark & \cmark & \cmark & \xmark & \cmark & \cmark & \cmark\\[0.2ex]
 \cline{2-9} \\[-2.5ex]
  & \GTNSS~\cite{GTN} & \cmark & \cmark & \cmark & \xmark & \cmark & \xmark & \cmark\\[0.2ex]
 \cline{1-9} \\[-2.5ex]
 -- & \BAGNNSS (ours) & \cmark & \cmark & \cmark & \cmark & \cmark & \cmark & \cmark\\[0.2ex] 
 \cline{1-9} 
 \end{tabularx}
 \begin{tablenotes}
\item $\dagger$ Personalized graph attention learns attention using the local graph neighborhood instead of the global graph context. 
\item *Primary baseline models
\item \textbf{Notation:} [A]: Does not require HoGs; [B]: Does not require a dynamic graph; [C]: Learns attention; [D]: Bi-level attention; [E] Personalized attention; [F]: Does not require meta paths; [G]: Transformer-inspired; (1): Non-GNN-based KGE models for HeGs; (2): GNNs for HoGs; (3A): Non-\TransformernospaceSS\textnormal{-}based GNNs for HeGs; (3B): \TransformernospaceSS\textnormal{-}based GNNs for HeGs
\end{tablenotes}
\vspace{-6mm}
\end{table}

\vspace{-2mm}
\subsection{GNNs for Homogeneous Graphs}
\vspace{-1mm}
Successful models in this category, like \GAT~\cite{GAT}, use attention-based neural architectures for learning representations.

% \vspace{-3.5mm}
\paragraph*{\textmd{\textbf{\normalsize Graph Attention Networks}}} \GAT~\cite{GAT} are additive attention-based GNNs that effectively leverage graph structure and sparsity to compute a node's attention. \GAT models, however, are limited to HoGs and cannot handle HeGs which contain different relations that may have varying levels of importance for different  nodes.

\vspace{-2mm}
\subsection{GNNs for Heterogeneous Graphs}
% \vspace{-1mm}
Successful models (1) leverage different graph relations, like \RGCN, (2) learn bi-level attention, like \HANnospace\textnormal, and (3) learn multiplicative attention, like \Transformernospace\textnormal{-based} NNs.

% \vspace{-4mm}
\paragraph*{\textmd{\textbf{\normalsize Relational Graph Convolutional Networks}}}
\RGCNnospace\textnormal{s}~\cite{rgcn} extend \GCNnospace\textnormal{s} and \GAT, which operate on local graph neighborhoods of HoGs, to operate on multi-relational graphs by distinguishing nodes by relation type. \RGCNnospace\textnormal{s}, however, treat all relation-specific nodes as equally important. Further, \RGCNnospace\textnormal{s} do not utilize graph attention as they are limited to directly learning from weight parameters.

% \vspace{-3.5mm}
\paragraph*{\textmd{\textbf{\normalsize Heterogeneous Graph Attention Networks}}}
To address limitations of the above models, \HAN integrates bi-level attention, which learns node- and relation-level attention, with GNNs to learn node embeddings. However, \HAN uses a global learnable weight vector lacking local inter-relation comparison. Besides, \HAN uses pre-defined metapaths which are computationally expensive to design and compute, and 
result in sub-optimal graph components learned by the model.

% \vspace{-3.5mm}
\paragraph*{\textmd{\textbf{\normalsize Transformer}}}
\Transformer models~\cite{Transformer}, although successful in natural language processing for small text sequences, have limitations for multi-relational and multi-entity HeGs. This is because \Transformer attends to all other tokens in the sequence, making it infeasible for large-scale input. While recent works extend \Transformernospace\textnormal{-like} attention to other graph domains, they have limitations, shown in Table~\ref{table:model-limitations}.

\vspace{-1mm}
\section{BA-GNN Architecture}
\label{BR-GCN-architecture}
\vspace{-1mm}
We design a general bi-level attention framework for computing hierarchical attention and then discuss \BAGNN's architecture. \normalsize{Source code and data are at:
\textcolor{black}{\url{https://github.com/roshnigiyer/BA-GNN}}. $\mathrm{READ.md}$ details dataset properties, data splits, and hyperparameters of \BAGNN models.}

\vspace{-1mm}
\subsection{General Bi-Level Attention Framework} 
\vspace{-1mm}
Bi-level attention in HeGs incorporates interactions between relation-specific nodes for learning lower level attention which informs the higher level attention that captures inter-relation interactions. In this way, bi-level attention jointly attends to node-node, relation-relation and node-relation interactions to collectively produce a representative node embedding. Uni-level attention models omit these critical graph interactions and ability for the two-levels of attention to jointly inform each other. Eq.~\ref{eqn:general_bi-level_attn} describes the general bi-level attention framework to compute embeddings for each $v_{i}$ in the HeG: 
% \yba{$\big( \Big( \bigg( \Bigg(	edge (i,j)$}
% \vspace{-1mm}
\begin{align}
\label{eqn:general_bi-level_attn}
    \Scale[0.9]{\widetilde{\mathbf{h}}_{i}^{(l+1)}} &= \Scale[0.9]{\mathrm{\mathbf{HigherAtt}}\big(\mathrm{\mathbf{LowerAtt}}(\cdot), \mathcal{R}, \widetilde{\mathbf{h}}^{(l)}\big)} \\
    &= \Scale[0.9]{\mathrm{AGG}\Bigg(\bigg\{\mathbf{f}_{\psi}\Big(\big\{\mathrm{\mathbf{LowerAtt}(\cdot)} \big| r \in \mathcal{R}, \widetilde{\mathbf{h}}^{(l)}\big\}\Big)\bigg| r \in \mathcal{R}\bigg\}\Bigg)}, \nonumber
\end{align}
% \vspace{-3mm}
\begin{gather}
\label{eqn:general_bi-level_attn-lower}
    \Scale[0.9]{\mathrm{\mathbf{LowerAtt}}(\cdot)} =
    \Scale[0.9]{\mathrm{AGG}\Big(\big\{\mathbf{g}_{\gamma}(e_{i,j}|r, \widetilde{\mathbf{h}}^{(l)})\big|v_{j} \in N_{i}^{r}\big\}\Big)}, \\
    \Scale[0.9]{\mathcal{R}} = \Scale[0.9]{\{\Gamma(e_{i,j}) | v_{j} \in \mathcal{V}\}},
    \vspace{-2mm}
\end{gather}
where ${\mathbf{h}_{i}}$ and $\widetilde{\mathbf{h}}_{i}$ are the initial and projected node features respectively, $\mathcal{R}_{i}$ is the relation set on the edge of $v_{i}$, and $\mathbf{g}_{\gamma}(\cdot)$ is a vector-output function of the node-level attention, $\gamma$, that provides a relation-specific embedding summary using learned node representations from the previous layer, $\widetilde{\mathbf{h}}^{(l)}$, which is aggregated, $\mathrm{AGG}(\cdot)$, over edges $e_{i,j}$ in the relation-specific neighborhood context of $v_{j} \in N_{i}^{r}$. $\mathbf{f}_{\psi}(\cdot)$ is a vector-output function of the relation-level attention, $\psi$, that are attended relation-specific local context embeddings, $\mathrm{\mathbf{LowerAtt}}(\cdot)$, which are aggregated over relations in the neighborhood context to form the layer's final node embedding, $\widetilde{\mathbf{h}}_{i}^{(l+1)}$. 

\begin{figure}[htp]
    \centering
    \includegraphics[width=9cm]{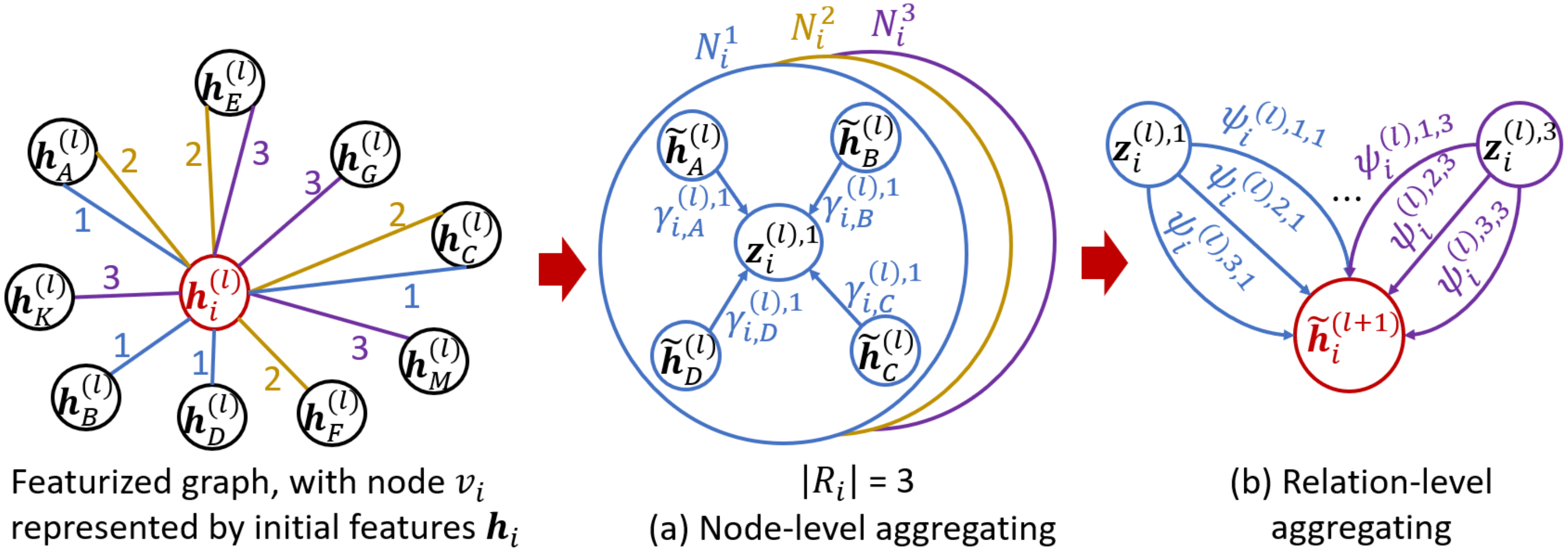}
    \caption{\textmd{\small{Bi-level attention visualization. (a) Node-level aggregating: A node’s features is a weighted combination of its prior layer's relation-specific embeddings, $\mathbf{z}_{i}^{r}$. (b) Relation-level aggregating:  Relation-level attention is learned via multiplicative attention using neighborhood relational similarity to determine relative relation importance.}}}
    \label{fig:2l-attention}
    % \vspace{-2mm}
\end{figure}

In Sections~\ref{subsection:node-level-att} and~\ref{subsection:relation-level-att}, we propose a novel semi-supervised attention-based \GCN model, \BAGNN, for multi-relational and multi-entity HeGs. \BAGNN performs attention aggregation on node and \textit{relation} levels, rather than nodes and edges. For node-level attention, attention is placed on the \textit{edges} of neighbor nodes. For relation-level attention, attention is placed on the relations, which are formed by \textit{grouping} edges by relation type. In this way, our model uses a hierarchical attention mechanism. The higher-order graph considers relation-type-specific edge groups, and the lower-order graph considers nodes in their local contexts. Figure~\ref{fig:2l-attention} summarizes \BAGNN's attention mechanism. \BAGNN models use $L$ stacked layers, each of which is defined through Eq.~\ref{eqn:general_bi-level_attn}, and for further efficiency, all nodes and relations are restricted to their neighborhoods. Model input can be chosen as pre-defined features or as a unique one-hot vector for each node.

\vspace{-4mm}

\vspace{-5mm}
\subsection{Node-level Attention}
\label{subsection:node-level-att}
\vspace{-0.5mm}
Node-level attention distinguishes different roles of nodes in the neighborhood context for learning relation-specific node embeddings. As node-level attentions are target-node-specific, they are different for different target nodes. Our model learns node embeddings such that it intrinsically captures graph attributes and structure in its neighborhood. In HeGs, neighbor nodes may belong to different feature spaces due to different node types, so we utilize an entity-specific type transformation matrix, $\boldsymbol{\mathcal{T}}_{\Lambda(v_{i})}$, to project all node features to the same space through $\widetilde{\mathbf{h}}_{i} = \boldsymbol{\mathcal{T}}_{\Lambda(v_{i})} \cdot \mathbf{h}_{i}$. $\boldsymbol{\mathcal{T}}_{\Lambda(v_{i})} \in \mathbb{R}^{d \times |\mathcal{R}_{i}|}, \mathbf{h}_{i} \in \mathbb{R}^{|\mathcal{R}_{i}|}$ if the initial features are chosen to be a one-hot vector of dimension $d$, 
% \vspace{-2mm}
% \begin{equation}
%     \label{eqn:projection}
%     \Scale[0.9]{
%     \widetilde{\mathbf{h}}_{i} = \boldsymbol{\mathcal{T}}_{\Lambda(v_{i})}} \cdot \mathbf{h}_{i},
%     % \vspace{-2mm}
% \end{equation}
where $\widetilde{\mathbf{h}}_{i}$ is continuously updated to learn the final embedding.   

\BAGNN's node-level attention uses additive attention inspired by \GAT, discussed in Section~\ref{related-work}, but overcomes \GAT's limitation by extending the attention to HeGs. \GAT performs projections that do not consider the different relation types in the HeG. We address this by using a learnable relation-specific attention vector, $\textbf{a}_{r}^{(l)} \in \mathbb{R}^{2d}$. For a specific relation $r$, the attention is shared for all node pairs, so that each node is influenced by its neighborhood context. The attention is also asymmetric since the importance of $v_{j}$ to $v_{i}$ may be different from the importance of $v_{i}$ to $v_{j}$. We compute relation-specific node-level attention at layer $l$ as follows, with a $\mathrm{softmax}(\cdot)$ activation applied to normalize each node-pair attention weight and where $v_j, v_k \in N_{i}^{r}$ and $\mathrm{\mathbf{x}}^{{T}^{(l)}}$ is transpose of $\mathrm{\mathbf{x}}$ at layer $l$: $\gamma_{i,j}^{(l),r} = \Scale[1.15]{\frac{\mathrm{exp}\Big(\mathrm{LeakyReLU}\big(\mathbf{a}_{r}^{T^{(l)}}\left[\widetilde{\mathbf{h}}_{i}^{(l)}\big|\big| \widetilde{\mathbf{h}}_{j}^{(l)}\right]\big)\Big)}{\sum_{v_{k} \in N_{i}^{r}}\mathrm{exp}\Big(\mathrm{LeakyReLU}\big(\mathbf{a}_{r}^{T^{(l)}}\left[\widetilde{\mathbf{h}}_{i}^{(l)}\big|\big| \widetilde{\mathbf{h}}_{k}^{(l)}\right]\big)\Big)}},$
% \vspace{-2mm}
% \begin{equation}
% \label{eqn:BR-GCN-node-level-attention}
% \Scale[0.95]{
% \gamma_{i,j}^{(l),r} =} \Scale[1.15]{\frac{\mathrm{exp}\Big(\mathrm{LeakyReLU}\big(\mathbf{a}_{r}^{T^{(l)}}\left[\widetilde{\mathbf{h}}_{i}^{(l)}\big|\big| \widetilde{\mathbf{h}}_{j}^{(l)}\right]\big)\Big)}{\sum_{v_{k} \in N_{i}^{r}}\mathrm{exp}\Big(\mathrm{LeakyReLU}\big(\mathbf{a}_{r}^{T^{(l)}}\left[\widetilde{\mathbf{h}}_{i}^{(l)}\big|\big| \widetilde{\mathbf{h}}_{k}^{(l)}\right]\big)\Big)}},
% \vspace{-2mm}
% \end{equation}
where $\textbf{a}_{r}^{(l)}$ attends over the concatenated, $||$, node features of $v_{i}$ and $v_{j}$ with applied $\mathrm{LeakyReLU}(\cdot)$ and $\mathrm{softmax}(\cdot)$ activations. By restricting the attention to within the relation-specific local context of nodes, sparsity structural information is injected into the model through adjacency-masked attention layers. 

Node $v_{i}$'s relation-specific embedding, $\textbf{z}_{i}^{(l),r}$, can then be learned with $\mathrm{AGG}(\cdot)$ from Eq.~\ref{eqn:general_bi-level_attn} being a weighted summation of the neighbor’s projected features as follows:
\vspace{-1mm}
\begin{align}
    \label{eqn:BR-GCN-semantic-specific-embedding}
    \Scale[0.95]{\mathbf{z}_{i}^{(l),r}} &= \Scale[0.9]{\mathrm{\mathbf{LowerAtt}}(\cdot)} \\ 
    &= \Scale[0.9]{\mathrm{AGG}\Big(\big\{\mathbf{g}_{\gamma}(e_{i,j}|r, \widetilde{\mathbf{h}}^{(l)})\big|v_{j} \in N_{i}^{r}\big\}\Big)} \nonumber \\
    &= \Scale[0.9]{\sum_{v_{j} \in N_{i}^{r}}\left[\mathbf{g}_{\gamma}( e_{i,j}|r, \widetilde{\mathbf{h}}^{(l)})\right]} \nonumber \\
    &= \Scale[0.9]{\sum_{v_{j} \in N_{i}^{r}}\left[\gamma_{i,j}^{(l),r}\widetilde{\mathbf{h}}_{j}^{(l)}\right]}, \nonumber
    \vspace{-4mm}
\end{align}
where $\mathbf{z}_{i}^{(l),r}$ provides a summary of relation $r$ for $v_{i}$ at layer $l$. We also add skip-connections for corresponding $\mathrm{\mathbf{LowerAtt}}(\cdot)$ from the previous layer, $l-1$, to preserve learned node-level representations as the depth of the NN is extended.   

\vspace{-1.5mm}
\subsection{Relation-level Attention}
\label{subsection:relation-level-att}
Relation-level attention distinguishes roles of different relations in the neighborhood context for learning more comprehensive node embeddings. In HeGs, different relations may play different roles of importance for $v_{i}$, in addition to $v_{i}$'s relation-specific neighbor nodes. So, we learn relation-level attention to better fuse $v_{i}$'s relation-specific node embeddings. One could design a simple node-relation attention mechanism to encode the effect of relations between nodes, but this would fail to capture relation-relation dependencies hidden in HeGs.  

We address \Transformer's inefficiency for large-scale HeGs through an approximation technique by sampling relation-specific node embeddings from the local graph context. Further, instead of using the same single set of projections for all words, we enable each relation-specific embedding to learn a distinct set of personalized projection weights, while maximizing parameter sharing. This technique captures unique relation-dependent characteristics such that each relation-specific node embedding is also influenced by its local relation context.

Node $v_{i}$'s relation-specific \Transformernospace\textnormal{-based} query $\textbf{q}_{r, i}^{(l)}$, key $\textbf{k}_{r, i}^{(l)}$, and value $\textbf{v}_{r, i}^{(l)}$ vectors are computed as follows: 
$\textbf{q}_{r, i}^{(l)}; \textbf{k}_{r, i}^{(l)}; \textbf{v}_{r, i}^{(l)} = \textbf{W}_{1,r}\textbf{z}_{i}^{(l),r}; \textbf{W}_{2,r}\textbf{z}_{i}^{(l),r}; \textbf{W}_{3,r}\textbf{z}_{i}^{(l),r},$
% \vspace{-1mm}
% \begin{equation}
% \label{eqn:BR-GCN-Transformer-QKV}
% \Scale[0.95]{
% \textbf{q}_{r, i}^{(l)}; \textbf{k}_{r, i}^{(l)}; \textbf{v}_{r, i}^{(l)} = \textbf{W}_{1,r}\textbf{z}_{i}^{(l),r}; \textbf{W}_{2,r}\textbf{z}_{i}^{(l),r}; \textbf{W}_{3,r}\textbf{z}_{i}^{(l),r}},
% \vspace{-1mm}
% \end{equation}
such that $\textbf{z}_{i}^{(l),r}$ is projected onto the learnable weight matrices \\ $\textbf{W}_{1,r}, \textbf{W}_{2,r}, \textbf{W}_{3,r} \in \mathbb{R}^{d \times d}$. The relation-level attention for relations $(r, r^{\prime})$ are computed by iterating over all possible relation pairs in the neighborhood context, $r,r^{\prime} \in \mathcal{R}_{i}$, where $\mathcal{R}_{i} = \{\Gamma(e_{i,j})|v_{j} \in \mathcal{V}_{i}\}$. The importance of relation $r^{\prime}$ of node $v_{i}$ is as follows, with relation similarity being captured through $\Scale[0.95]{\psi_{i}^{(l),r,r^{\prime}}} = \Scale[0.95]{\mathrm{softmax}(\mathbf{q}_{r, i}^{T^{(l)}}\mathbf{k}_{r^{\prime}, i}^{^{(l)}})},$
% \vspace{-1mm}
% \begin{equation}
%     \label{eqn:BR-GCN-semantic-level}
%     \Scale[0.95]{\psi_{i}^{(l),r,r^{\prime}}} = \Scale[0.95]{\mathrm{softmax}(\mathbf{q}_{r, i}^{T^{(l)}}\mathbf{k}_{r^{\prime}, i}^{^{(l)}})},
%     % \vspace{-1mm}
% \end{equation}
where the more similar $r^{\prime}$ is to $r$, the greater the attention weights of $r^{\prime}$, which results in more contribution of $r^{\prime}$'s embedding to $v_{i}$'s final embedding. A $\mathrm{softmax}(\cdot)$ activation is then applied to normalize each relation pair's attention weight.

A node's relation-specific embedding is then informed by a weighted summation of its similarity to other local context relations, $\psi_{i}^{(l),r, r^{\prime}}$. To reduce information loss, we add a self-connection of a special relation type per node, which is projected onto $\mathbf{W}_{i}$, and aggregated to the attended relation-specific embedding, $\psi_{i}^{(l),r,r^{\prime}}\mathbf{v}_{r^{\prime},i}^{(l)}$. Lastly, a $\mathrm{ReLU}(\cdot)$ activation is applied to get the overall relation-specific embedding,     $\boldsymbol{\delta}_{i}^{(l),r} = \mathrm{ReLU}(\sum_{r^{\prime} \in \mathcal{R}_{i}}\psi_{i}^{(l),r, r^{\prime}}\mathbf{v}_{r^{\prime},i}^{(l)} + \mathbf{W}_{i}\widetilde{\mathbf{h}}_{i}^{(l)})$.
% \vspace{-1mm}
% \begin{equation}
%     \Scale[0.9]{
%     \boldsymbol{\delta}_{i}^{(l),r} = \mathrm{ReLU}(\sum_{r^{\prime} \in \mathcal{R}_{i}}\psi_{i}^{(l),r, r^{\prime}}\mathbf{v}_{r^{\prime},i}^{(l)} + \mathbf{W}_{i}\widetilde{\mathbf{h}}_{i}^{(l)})}.
% \end{equation}

Node $v_{i}$'s final embedding is learned with $\mathrm{AGG}(\cdot)$ in Eq.~\ref{eqn:general_bi-level_attn} being a summation of all attended relation-specific embeddings through iteration of neighborhood relations, $r \in R_{i}$. We also apply multi-head attention of $S = \{1,...,K\}$ heads to allow \BAGNN to jointly attend to different representation subspaces of nodes/relations, with aggregation, $\mathrm{AGG}(\cdot)$, via averaging:
\vspace{-1mm}
\begin{align}
\label{eqn:BR-GCN-final-embedding}
\Scale[0.9]{\widetilde{\mathbf{h}}_{i}^{(l+1)}} &= \Scale[0.9]{\mathrm{\mathbf{HigherAtt}}\big(\mathrm{\mathbf{LowerAtt}}(\cdot), \mathcal{R}, \widetilde{\mathbf{h}}^{(l)}\big)} \\
&= \Scale[0.9]{\Scale[1.2]{\frac{1}{K}}\sum_{k=1}^{K}\sum_{r \in \mathcal{R}}\bigg[\mathbf{f}_{\psi}\Big(\big\{ \mathrm{\mathbf{LowerAtt}(\cdot)}\big|r \in \mathcal{R}, \widetilde{\mathbf{h}}^{(l)}\big\}\Big)\bigg]} \nonumber \\
&= \Scale[0.9]{\Scale[1.2]{\frac{1}{K}}\sum_{k=1}^{K}\sum_{r \in \mathcal{R}_{i}}\bigg[\mathbf{f}_{\psi}\Big(\big\{ \mathrm{\mathbf{LowerAtt}(\cdot)}\big|r \in \mathcal{R}_{i}, \widetilde{\mathbf{h}}^{(l)}\big\}\Big)\bigg]} \nonumber \\
&= \Scale[0.9]{\mathrm{AGG}\Big(\big\{\sum_{r \in \mathcal{R}_{i}} \left[\boldsymbol{\delta}_{i}^{(l),r}\right]}\big| k \in S \} \Big). \nonumber
\vspace{-1mm}
\end{align}
We also add skip-connections for corresponding $\mathrm{\mathbf{HigherAtt}}(\cdot)$ from the previous layer, $l-1$, to preserve learned higher-order relation-level representations as depth of the NN is extended. 

The final representation of a node at layer $(l + 1)$ is:
\vspace{-2mm}
\begin{multline}
    \Scale[0.9]{
    \widetilde{\mathbf{h}}_{i}^{(l+1)} = \mathrm{AGG}\bigg(\Big\{\sum_{r \in \mathcal{R}_{i}} \mathrm{ReLU}\big(}\\
    \Scale[0.9]{\sum_{r^{\prime} \in \mathcal{R}_{i}}\mathrm{softmax}(\mathbf{q}_{r, i}^{T^{(l)}}\mathbf{k}_{r^{\prime}, i}^{^{(l)}})\mathbf{v}_{r^{\prime}, i}^{^{(l)}} + \mathbf{W}_{i}\widetilde{\mathbf{h}}_{i}^{(l)}\big)} \Big| k \in S \Big\}\bigg).
\end{multline}

\vspace{-4mm}
\subsection{Analysis of Proposed Attention}
We use multiplicative attention at the relation level, instead of additive attention, because learning attention through a concatenation of features does not compute feature similarity which inner product operations capture. Since relation features are characterized by single attribute relation types, a relation's scaling can be directly determined by its feature similarity to other relations in its neighborhood. This is unlike node-level attention, where node features may have several attributes, making it more difficult to learn latent similarity of nodes through direct feature comparison such as through inner product computation. Rigorous evaluation of mixed combinations of additive/multiplicative attention shown in experiments further support our bi-level attention combination choice.
\vspace{-3mm}
\section{Experiments}
\label{evaluation}
\vspace{-1mm}
In this section, we evaluate \BAGNN on seven large-scale heterogeneous datasets (HDs). We conduct experiments on node classification and link prediction using Pytorch Geometric and Deep Graph Library frameworks on an Nvidia Tesla V100 GPU cluster, and report model test accuracies. 

% \vspace{-3mm}
\paragraph*{\textmd{\textbf{\normalsize Datasets}}} We evaluate on benchmark Resource Description Framework (RDF) format datasets~\cite{RDF} for node classification: AIFB, MUTAG, BGS, and AM. For link prediction, we evaluate on FB15k~\cite{FB15k-237}, WN18~\cite{WIN18RR}, and FB15k-237~\cite{FB15k-237}.

\setul{}{0.6pt}
\definecolor{Gray}{gray}{0.9}
\newcommand{\mycc}{\cellcolor{Gray}}
\newcolumntype{g}{>{\columncolor{Gray}}c}

  \begin{table*}[t!]
  \scriptsize
  \caption{\textmd{\small{
  Node classification averaged accuracy (\%) over 10 model trains on four datasets. Best results per model group are bold-faced, with overall best results per dataset underscored. We compute test accuracy improvement of \BAGNN over best NNs, all NNs, and per NN by the difference of \BAGNNnospace\textnormal{'s} average accuracy and NNs per dataset column (e.g., rows 24-25), or the difference of \BAGNNnospace\textnormal{'s} average accuracy and each NN across datasets (col. 7). $\textsc{A}_\textsc{node, rel.}/\textsc{M}_\textsc{node, rel.}$ are \BAGNN additive/mutliplicative attention at node/relation levels.}}}
  \label{table:entity-classificaton-results}
  \centering
  \newcolumntype{Y}{>{\centering\arraybackslash}X}
  \begin{tabularx}{\linewidth}{cYccccg}
  \cline{1-7} \\[-2.3ex]
 Type & Model & AIFB & MUTAG & BGS & AM & \BAGNNSS (\% improve)\\ [0.5 ex]
\cline{1-7} \\[-2.3ex]
\multicolumn{1}{c|}{} & \multicolumn{1}{c|}{\TransESS~\cite{TransE}} & 66.89 $\pm$ 1.26 & 54.41 $\pm$ 0.73 & 58.46 $\pm$ 0.38 & 61.23 $\pm$ 0.57 & +[32.05, 36.29]\\[0.5ex]
 \multicolumn{1}{c|}{(1) Non-GNN-based} & \multicolumn{1}{c|}{\HolESS~\cite{HolE}} & 67.98 $\pm$ 0.42 & \textbf{63.17} $\pm$ 0.26 & 72.74 $\pm$ 0.22 & 65.25 $\pm$ 0.38 & +[22.01, 31.43]\\[0.5ex]
  \multicolumn{1}{c|}{KGE models for HeGs} & \multicolumn{1}{c|}{\DistMultSS~\cite{Distmult}} & 73.64 $\pm$ 0.10 & 62.06 $\pm$ 0.18 & 68.19 $\pm$ 0.13 & 69.12 $\pm$ 0.24 & +[25.3, 27.56]\\[0.5ex]
 \multicolumn{1}{c|}{} & \multicolumn{1}{c|}{\ComplExSS~\cite{ComplEx}} & \textbf{77.24} $\pm$ 0.15 & 62.38 $\pm$ 0.22 & \textbf{74.72} $\pm$ 0.18 & \textbf{72.39} $\pm$ 0.16 & +[20.03, 25.43]\\[0.5ex]
 \cline{1-7} \\[-2.3ex]
 \multicolumn{1}{c|}{(2) GNNs for} & \multicolumn{1}{c|}{\GCNSS~\cite{GCN}} & 91.99 $\pm$ 0.21 & \textbf{67.02} $\pm$ 0.08 & \textbf{78.74} $\pm$ 0.16 & 86.82 $\pm$ 0.60 & +[6.95, 20.79]\\[0.5ex]
 \multicolumn{1}{c|}{HoGs} & \multicolumn{1}{c|}{\GATSS~\cite{GAT}} & \textbf{92.50} $\pm$ 0.29 & 66.18 $\pm$ 0.00 & 77.93 $\pm$ 0.17 & \textbf{88.52} $\pm$ 1.65 & +[6.44, 21.63]\\[0.5ex]
 \cline{1-7} \\[-2.3ex]
 \multicolumn{1}{c|}{} & \multicolumn{1}{c|}{\metavecSS~\cite{metapath2vec}} & 89.52 $\pm$ 0.12 & 66.04 $\pm$ 0.27 & 78.34 $\pm$ 0.13 & 85.48 $\pm$ 0.11 & +[9.42, 21.77]\\[0.5ex]
 \multicolumn{1}{c|}{} & \multicolumn{1}{c|}{\HERecSS~\cite{HERec}} & 91.03 $\pm$ 0.15 & 66.96 $\pm$ 0.18 & 79.36 $\pm$ 0.25 & 85.98 $\pm$ 0.07 & +[7.91, 20.85]\\[0.5ex]
 \multicolumn{1}{c|}{} & \multicolumn{1}{c|}{\HINVecSS~\cite{HIN2Vec}} & 91.63 $\pm$ 0.17 & 66.29 $\pm$ 0.14 & 79.01 $\pm$ 0.12 & 86.22 $\pm$ 0.21 & +[7.31, 21.52] \\[0.5ex]
 \multicolumn{1}{c|}{(3A) Non-\TransformernospaceSS\textnormal{-}} & \multicolumn{1}{c|}{\HeGANSS~\cite{HeGAN}} & 92.33 $\pm$ 0.13 & 68.07 $\pm$ 0.08 & 81.60 $\pm$ 0.27 & 86.79 $\pm$ 0.14 & +[6.61, 19.74]\\[0.5ex]
 \multicolumn{1}{c|}{based GNNs for HeGs} & \multicolumn{1}{c|}{\TemporalGATSS~\cite{TemporalGAT}} & 93.42 $\pm$ 0.11 & 66.88 $\pm$ 0.24 & 79.14 $\pm$ 0.13 & 89.10 $\pm$ 0.13 & +[5.52, 20.93]\\[0.5ex]
 \multicolumn{1}{c|}{} & \multicolumn{1}{c|}{\HetGNNSS~\cite{HetGNN}} & 95.18 $\pm$ 0.16 & 75.64 $\pm$ 0.09 & 82.05 $\pm$ 0.25 & 89.67 $\pm$ 0.05 & +[3.76, 12.70]\\[0.5ex]
 \multicolumn{1}{c|}{} & \multicolumn{1}{c|}{\RGCNSS~\cite{rgcn}} & 95.31 $\pm$ 0.62 & 73.23 $\pm$ 0.48 & 83.10 $\pm$ 0.80 & 89.29 $\pm$ 0.35 & +[3.63, 14.58]\\[0.5ex]
 \multicolumn{1}{c|}{} & \multicolumn{1}{c|}{\HANSS~\cite{HAN}} & \textbf{96.25} $\pm$ 0.12 & \textbf{76.46} $\pm$ 0.07 & \textbf{86.84} $\pm$ 0.21 &
 \textbf{90.68} $\pm$ 0.23 & +[2.69, 11.35]\\[0.5ex]
 \cline{1-7} \\[-2.3ex]
 \multicolumn{1}{c|}{} & \multicolumn{1}{c|}{\DySATSS~\cite{DySAT}} & 92.64 $\pm$ 0.21 & 66.57 $\pm$ 0.05 & 78.02 $\pm$ 0.19 & 88.90 $\pm$ 1.05 & +[6.30, 21.24]\\[0.5ex]
 \multicolumn{1}{c|}{(3B) \TransformernospaceSS\textnormal{-based}} & \multicolumn{1}{c|}{\TGATSS~\cite{TGAT}} & 92.84 $\pm$ 0.14 & 67.19 $\pm$ 0.21 & 78.35 $\pm$ 0.15 & 89.43 $\pm$ 0.28 & +[6.10, 20.62]\\[0.5ex]
 \multicolumn{1}{c|}{GNNs for HeGs} & \multicolumn{1}{c|}{\HGTSS~\cite{HGT}} & 95.97 $\pm$ 0.15 & \textbf{76.84} $\pm$ 0.12 & \textbf{86.01} $\pm$ 0.18 & 90.33 $\pm$ 0.13 & +[2.97, 10.97]\\[0.5ex]
  \multicolumn{1}{c|}{} & \multicolumn{1}{c|}{\GTNSS~\cite{GTN}} & \textbf{96.04} $\pm$ 0.17 & 76.32 $\pm$ 0.12  & 85.38 $\pm$ 0.24 & \textbf{90.56} $\pm$ 0.10 & +[2.90, 11.49]\\[0.5ex]
 \cline{1-7} \\ [-2.3ex]
 \multicolumn{1}{c|}{} & \multicolumn{1}{c|}{\BAGNNnodeSS} & 95.46 $\pm$ 0.13 & 73.19 $\pm$ 0.25 & 84.23 $\pm$ 0.22 & 89.45 $\pm$ 0.02 & --\\[0.5ex]
 \multicolumn{1}{c|}{\BAGNNSS variants} & \multicolumn{1}{c|}{\BAGNNrelSS} & 95.28 $\pm$ 0.23 & 76.17 $\pm$ 0.22 & 85.43 $\pm$ 0.34 & 90.52 $\pm$ 0.18 & --\\[0.5ex]
%  \cline{2-7} \\[-2.3ex]
%  \cline{2-7} \\[-2.3ex]
 \multicolumn{1}{c|}{} & \multicolumn{1}{c|}{\BAGNNhanSS} & 96.30 $\pm$ 0.09 & 76.48 $\pm$ 0.04 & 86.80 $\pm$ 0.24 & 90.67 $\pm$ 0.35 & --\\[0.5ex] 
  \multicolumn{1}{c|}{} & \multicolumn{1}{c|}{\BAGNNSS ($\textsc{M}_\textsc{node}/\textsc{A}_\textsc{rel.}$)} & 96.02 $\pm$ 0.13 & 76.20 $\pm$ 0.09 & 86.90 $\pm$ 0.17 & 90.54 $\pm$ 0.11 & --\\ [0.5 ex]
 \multicolumn{1}{c|}{} & \multicolumn{1}{c|}{\BAGNNSS ($\textsc{A}_\textsc{node}/\textsc{A}_\textsc{rel.}$)} & 96.38 $\pm$ 0.14 & 76.61 $\pm$ 0.22 & 87.00 $\pm$ 0.09 & 90.73 $\pm$ 0.05 & --\\ [0.5 ex]
 \multicolumn{1}{c|}{} & \multicolumn{1}{c|}{\BAGNNSS ($\textsc{M}_\textsc{node}/\textsc{M}_\textsc{rel.}$)} & 96.44 $\pm$ 0.16 & 77.01 $\pm$ 0.07 & 86.92 $\pm$ 0.12 & 90.78 $\pm$ 0.08 & --\\ [0.5 ex]
 \multicolumn{1}{c|}{} & \multicolumn{1}{c|}{\BAGNNSS (ours)} & \ul{\textbf{98.94}} $\pm$ 0.13 & \ul{\textbf{87.81}} $\pm$ 0.11 & \ul{\textbf{94.75}} $\pm$ 0.08 & \ul{\textbf{96.68}} $\pm$ 0.14 & --\\ [0.5 ex]
 \rowcolor{Gray}
 \multicolumn{1}{c|}{\BAGNNSS (\% improve)} & \multicolumn{1}{c|}{against \textit{\textbf{best}} NNs, group 1-3} & +[2.69, 21.70] & +[10.97, 24.64] & +[7.91, 20.03] & +[6.00, 24.29] & --\\ [0.5 ex]
 \rowcolor{Gray}
 \multicolumn{1}{c|}{} & \multicolumn{1}{c|}{against \textit{\textbf{all}} NNs, group 1-3} & +[2.69, 32.05] & +[10.97, 33.40] & +[7.91, 36.29] & +[6.00, 35.45] & --\\ [0.5 ex]
 \cline{1-7}
 \end{tabularx}
 \vspace{-3mm}
\end{table*}

\vspace{-3mm}
\subsection{Node Classification}
Node classification is the semi-supervised classification of nodes to entity types. For evaluation consistency against primary baseline models, we implement \BAGNN with $L = 2$ and where the output of the final layer uses a $\mathrm{softmax}(\cdot)$ activation per node. Our model follows the same node classification evaluation procedure as~\cite{rgcn}, using cross-entropy loss with parameters learned from the Adam Optimizer. 

% \vspace{-3mm}
\paragraph*{\textmd{\textbf{\normalsize Baselines}}} Table~\ref{table:model-limitations} summarizes our baselines. To adapt the models to our problem setting of multi-relational, static HeGs, we made the following modifications. For \GAT~\cite{GAT} and \GCN~\cite{GCN}, we omit HeG relations. For \TemporalGAT~\cite{TemporalGAT}, we omit the temporal convolutional network used for temporal interactions. For \HetGNN~\cite{HetGNN}, we consider the neighbors to be the entire set of neighbor nodes and relations. For \DySAT~\cite{DySAT}, we omit temporal attention. For \TGAT~\cite{TGAT}, we omit functional time encoding. For \HGT~\cite{HGT}, we omit relative temporal encoding.
% \vspace{-3mm}

% \begin{itemize}
% % \setlength\itemsep{-0.4em}
%     \item \textbf{\GAT~\cite{GAT}} and \textbf{\GCN~\cite{GCN}}: We omit HeG relations.
    
%     \item \textbf{\TemporalGAT~\cite{TemporalGAT}}: We omit the temporal convolutional network used for temporal interactions.
    
%     \item \textbf{\HetGNN~\cite{HetGNN}}: We consider the neighbors to be the entire set of neighbor nodes and relations. 
    
%     \item \textbf{\DySAT~\cite{DySAT}}: We omit temporal attention.
    
%     \item \textbf{\TGAT~\cite{TGAT}}: We omit functional time encoding.
    
%     \item \textbf{\HGT~\cite{HGT}}: We omit relative temporal encoding. 

% \end{itemize}

% \vspace{-1mm}

\begin{table*}[t!]
\scriptsize
\caption{\small{Link prediction results for mean reciprocal rank (MRR), and Hits @ n metrics. For each group of models, the best results are bold-faced. The overall best results on each dataset are underscored.}}
\centering
\newcolumntype{Y}{>{\centering\arraybackslash}X}
\begin{tabularx}{\linewidth}{@{}Y@{\hskip8pt}c@{\hskip8pt}c@{\hskip8pt}c@{\hskip8pt}c@{\hskip8pt}c@{\hskip8pt}|c@{\hskip8pt}c@{\hskip8pt}c@{\hskip8pt}c@{\hskip8pt}c@{\hskip8pt}|c@{\hskip8pt}c@{\hskip8pt}c@{\hskip8pt}c@{\hskip8pt}c@{\hskip8pt}@{}}
\cline{1-16} \\ [-1.5ex]
      & \multicolumn{5}{c}{FB15k}                             & \multicolumn{5}{c}{WN18} & \multicolumn{5}{c}{FB15k-237}                            \\ \cmidrule(lr){2-6} \cmidrule(l){7-11} \cmidrule(l){12-16}
      & \multicolumn{2}{c}{MRR} & \multicolumn{3}{c}{Hits @} & \multicolumn{2}{c}{MRR} & \multicolumn{3}{c}{Hits @} & \multicolumn{2}{c}{MRR} & \multicolumn{3}{c}{Hits @} \\ \cmidrule(lr){2-3} \cmidrule(lr){4-6} \cmidrule(lr){7-8} \cmidrule(l){9-11} \cmidrule(l){12-13} \cmidrule(l){14-16}
Model & Raw      & Filtered     & 1       & 3      & 10    & Raw      & Filtered     & 1       & 3      & 10 & Raw      & Filtered     & 1       & 3      & 10      \\ \cline{1-16} \\[-2.4ex]
\RGCNSS & 0.251 & 0.651 & 0.541 & 0.736 & 0.825 & 0.553 & 0.814 & 0.686 & 0.928 & 0.955 & 0.158 & 0.248 & 0.153 & 0.258 & 0.414\\ [0.5ex]
\BAGNNSS & \textbf{0.261} & \textbf{0.702} & \textbf{0.601} & \textbf{0.778} & \textbf{0.857} & \textbf{0.590} & \textbf{0.820} & \textbf{0.698} & \textbf{0.945} & \textbf{0.959} & \textbf{0.195} & \textbf{0.260} & \textbf{0.160} & \textbf{0.268} & \textbf{0.468} \\ [0.5ex]
\cline{1-16} \\[-2.4ex]
\TransESS & 0.221 & 0.380 & 0.231 & 0.472 & 0.641 & 0.335 & 0.454 & 0.089 & 0.823 & 0.934 & 0.144 & 0.233 & 0.147 & 0.263 & 0.398 \\  [0.5ex]
\RGCNnospaceSS\textnormal{$_{T}$}  			& 0.252 & 0.651 & 0.543 & 0.738 & 0.828 & 0.554 & 0.815 & 0.681 & 0.928 & 0.956 & 0.161 & 0.258 & 0.159 & 0.274 & 0.421 \\ [0.5ex]
\BAGNNnospaceSS\textnormal{$_{T}$}  			& \textbf{0.264} & \textbf{0.700} & \textbf{0.649} & \textbf{0.781} & \textbf{0.858} & \textbf{0.593} & \textbf{0.822} & \textbf{0.692} & \textbf{0.943} & \textbf{0.960} & \ul{\textbf{0.200}} & \ul{\textbf{0.268}} & \ul{\textbf{0.168}} & \ul{\textbf{0.275}} & \ul{\textbf{0.493}} \\ [0.5ex]
\cline{1-16} \\[-2.4ex]
\HolESS & 0.232 & 0.524 & 0.402 & 0.613 & 0.739 & 0.616 & 0.938 & 0.930 & 0.945 & 0.949 & 0.124 & 0.222 & 0.133 & 0.253 & 0.391 \\ [0.5ex]
\RGCNnospaceSS\textnormal{$_{H}$}  			& 0.257 & 0.659 & 0.556 & 0.744 & 0.839 & 0.667 & 0.937 & 0.935 & 0.951 & 0.966 & 0.159 & 0.257 & 0.156 & \textbf{0.272} & 0.420 \\ [0.5ex]
\BAGNNnospaceSS\textnormal{$_{H}$}  			& \textbf{0.268} & \textbf{0.720} & \textbf{0.670} & \textbf{0.787} & \textbf{0.860} & \textbf{0.670} & \textbf{0.940} & \textbf{0.942} & \textbf{0.955} & \textbf{0.979} & \textbf{0.194} & \textbf{0.266} & \textbf{0.161} & \textbf{0.272} & \textbf{0.488} \\  [0.5ex]
\cline{1-16} \\[-2.4ex]
\DistMultSS & 0.248 & 0.634 & 0.522 & 0.718 & 0.814 & 0.526 & 0.813 & 0.701 & 0.921 & 0.943 & 0.100 & 0.191 & 0.106 & 0.207 & 0.376 \\ [0.5ex]
\RGCNnospaceSS\textnormal{$_{D}$}			& 0.262 & 0.696 & 0.601 & 0.760 & 0.842 & 0.561 & 0.819 & 0.697 & 0.929 & 0.964 & 0.156 & 0.249 & 0.151 & 0.264 & 0.417  \\ [0.5ex]
\BAGNNnospaceSS\textnormal{$_{D}$}  			& \textbf{0.272} & \textbf{0.745} & \textbf{0.688} & \textbf{0.792} & \textbf{0.868} & \textbf{0.600} & \textbf{0.825} & \textbf{0.705} & \textbf{0.934} & \textbf{0.977} & \textbf{0.190} & \textbf{0.251} & \textbf{0.155} & \textbf{0.268} & \textbf{0.483}\\ [0.5ex]
\cline{1-16} \\[-2.4ex]
\ComplExSS & 0.242 & 0.692 & 0.599 & 0.759 & 0.840 & 0.587 & 0.941 & 0.936 & 0.945 & 0.947 & 0.109 & 0.201 & 0.112 & 0.213 & 0.388 \\ [0.5ex]
\RGCNnospaceSS\textnormal{$_{C}$}  			& 0.260 & 0.712 & 0.629 & 0.771 & 0.845 & 0.615 & 0.953 & 0.937 & 0.947 & 0.965 & 0.158 & 0.255 & 0.152 & 0.268 & 0.419 \\ [0.5ex]
\BAGNNnospaceSS\textnormal{$_{C}$}  			& \ul{\textbf{0.278}} & \ul{\textbf{0.788}} & \ul{\textbf{0.731}} & \ul{\textbf{0.867}} & \ul{\textbf{0.880}} & \ul{\textbf{0.671}} & \ul{\textbf{0.978}} & \ul{\textbf{0.981}} & \ul{\textbf{0.988}} & \ul{\textbf{0.992}} & \textbf{0.170} & \textbf{0.262} & \textbf{0.159} & \textbf{0.270} & \textbf{0.485}\\ [0.5ex]
\cline{1-16}
\end{tabularx}
\label{table:link-results}
\vspace{-4mm}
\end{table*}

% \vspace{-6.5mm}
\newcommand{\fbseries}{\unskip\setBold\aftergroup\unsetBold\aftergroup\ignorespaces}
\newcommand{\setBoldness}[1]{\def\fake@bold{#1}}
\paragraph*{\textmd{\textbf{\normalsize Results}}} 
Experiment results are in Table~\ref{table:entity-classificaton-results}.
Results show \BAGNN significantly and consistently outperforms all baselines for all tasks on all datasets. For example, on AIFB, MUTAG, BGS, and AM, against the most competitive NNs per category, \BAGNN achieves relative performance gains of up to 22\%, 25\%, 20\%, and 24\% respectively, and overall performance gains of up to 32\%, 33\%, 36\%, and 35\% respectively. Further, the Welch t-test of unequal variance shows that \BAGNN's relative performance compared to each model per dataset is statistically significant to be greater, with $\mathrm{p}$-$\mathrm{value} < 0.001$.

% \vspace{-3mm}
\paragraph*{\textmd{\textbf{\normalsize Ablation Studies}}}
To further analyze \BAGNN's bi-level attention, we design the following \BAGNN variant models: \BAGNNnode, a uni-level attention model using \BAGNN's node-level attention; \BAGNNrel, a uni-level attention model using \BAGNN's relation-level attention; \BAGNNhan, a hybrid model using \BAGNN's node-level attention and \HAN's relation-level attention.

% \vspace{-3mm}

% \begin{itemize}
% % \setlength\itemsep{-0.4em}

% \item \textbf{\BAGNNnode}, which is a uni-level attention model using \BAGNN's node-level attention

% \item \textbf{\BAGNNrel}, which is a uni-level attention model using \BAGNN's relation-level attention

% \item \textbf{\BAGNNhan}, a hybrid model using \HAN's relation-level attention. 
% \end{itemize}

The bi-level attention models outperform the uni-level attention models, shown in Table~\ref{table:entity-classificaton-results}. Further, simply comparing \BAGNN's higher-order relation-level attention with \HAN's, shows a significant relative performance gain on all datasets. For example on MUTAG, the performance gain is nearly $11.30\%$ with the replacement of only the relation-level attention. This indicates that \BAGNN's relation-level attention is more effective across the different data domains and that its personalized attention to local graph contexts yields performance gain.

\vspace{-4mm}
\subsection{Link Prediction}
\label{sec:eval-link-pred}
\vspace{-1.5mm}
Link prediction involves assigning confidence scores to HeG triples to determine how likely predicted edges belong to true relations. Our models follow the same evaluation framework as ~\cite{TransE} and~\cite{rgcn} using negative sampling and cross-entropy loss with parameters learned from the Adam Optimizer. We use evaluation metrics of mean reciprocal rank (MRR) and Hits @ n, in raw and filtered settings. The same number of negative samples, $w = 1$, are used to make datasets comparable.

% \vspace{-3mm}
\paragraph*{\textmd{\textbf{\normalsize Baselines}}}
We evaluate standalone GNNs (\BAGNN, \RGCN), KGE models, and GNN+KGE autoencoder models using the same setup procedure as~\cite{TransE} and~\cite{rgcn}.
The autoencoder models include: \BAGNNnospace\textnormal{$_{x}$} and \RGCNnospace\textnormal{$_{y}$} where $x$, $y$ are \TransE ($T$), \HolE ($H$), \DistMult ($D$), and \ComplEx ($C$). 

% \vspace{-6mm}
\paragraph*{\textmd{\textbf{\normalsize Results and Ablation Studies}}}
Experiment results are in Table~\ref{table:link-results}. Results show that the best \BAGNN models outperform \RGCN models on all datasets for all metrics of both tasks of MRR and Hits @ n = 1, 3, 10. We observe that \BAGNN outperforms \RGCN when comparing standalone models. Further, results show that autoencoder models outperform each of GNN and KGE standalone models, showing that GNNs and KGE models can each be benefited by their joint learning. Results also show that \BAGNN autoencoders outperform \RGCN autoencoders on all datasets for all tasks and metrics.

\vspace{-1mm}
\subsection{Case Study}
\begin{figure}[!htp]
\vspace{-2mm}
    \subfloat[\centering BA-GNN’s attention-induced
graph structure for \HANSS, \BAGNNnodeSS, \BAGNNrelSS, \BAGNNSS]
    {
    {\includegraphics[width=2.2cm]{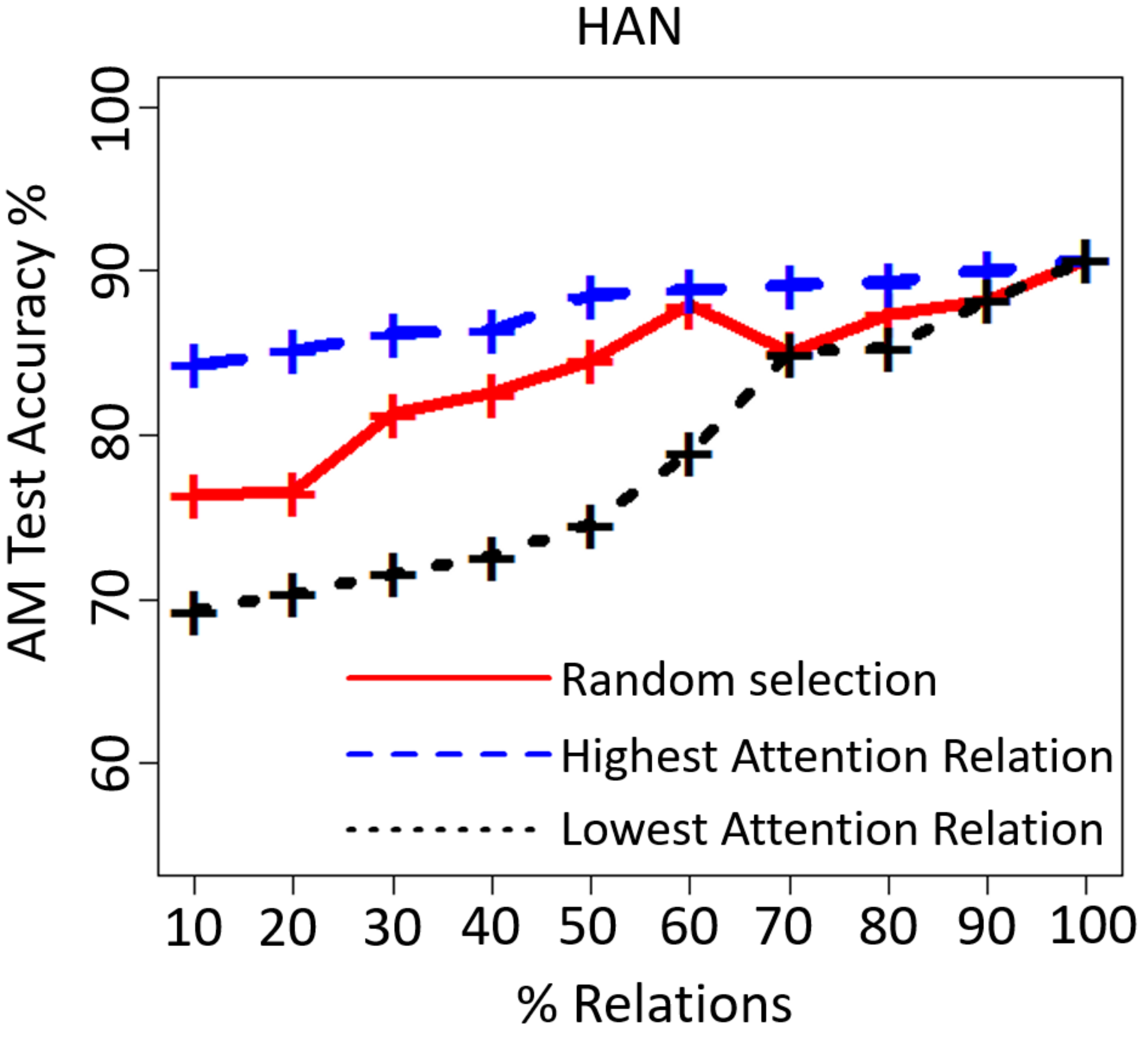}}
    {\includegraphics[width=2.2cm]{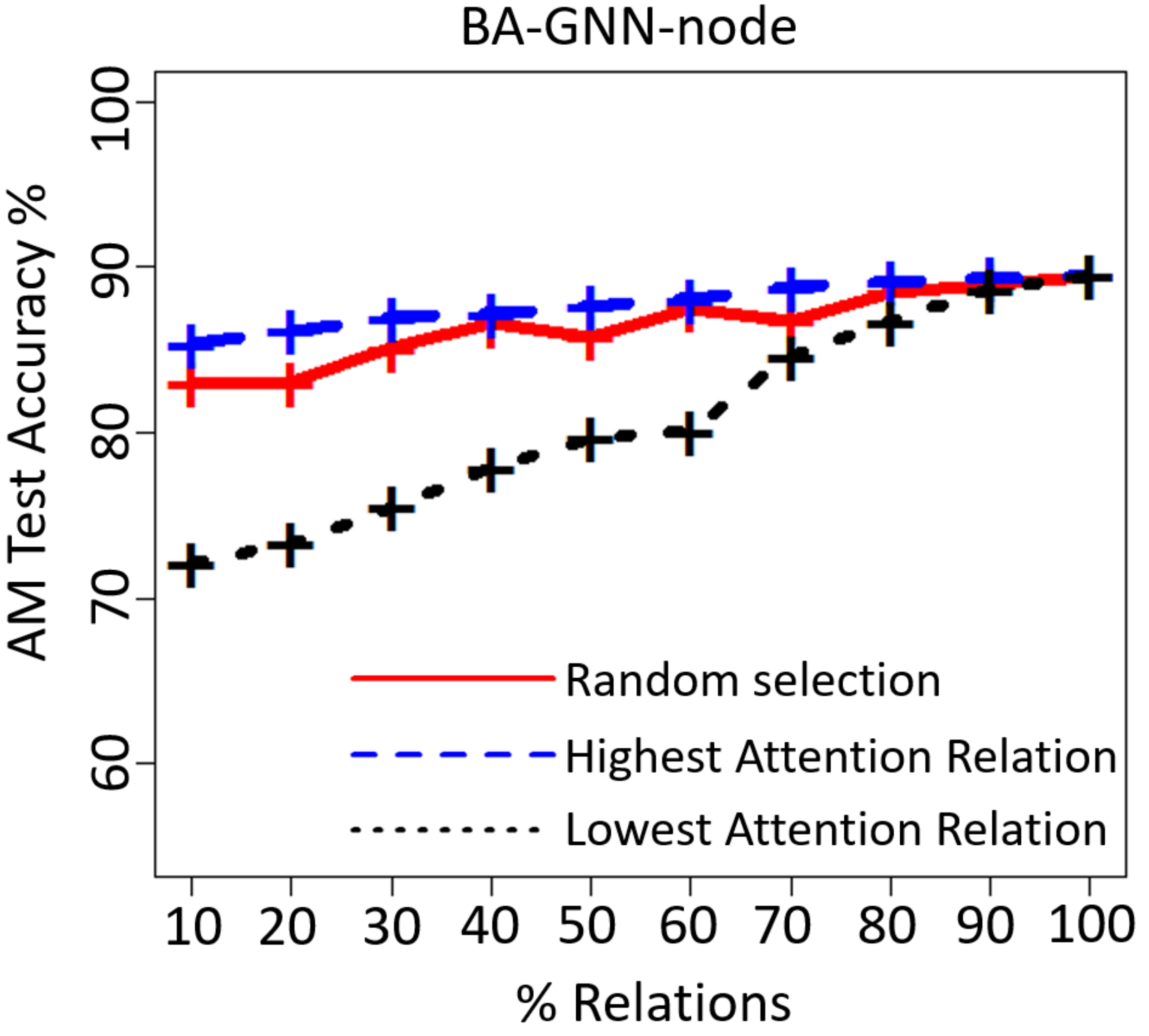}}
    {\includegraphics[width=2.2cm]{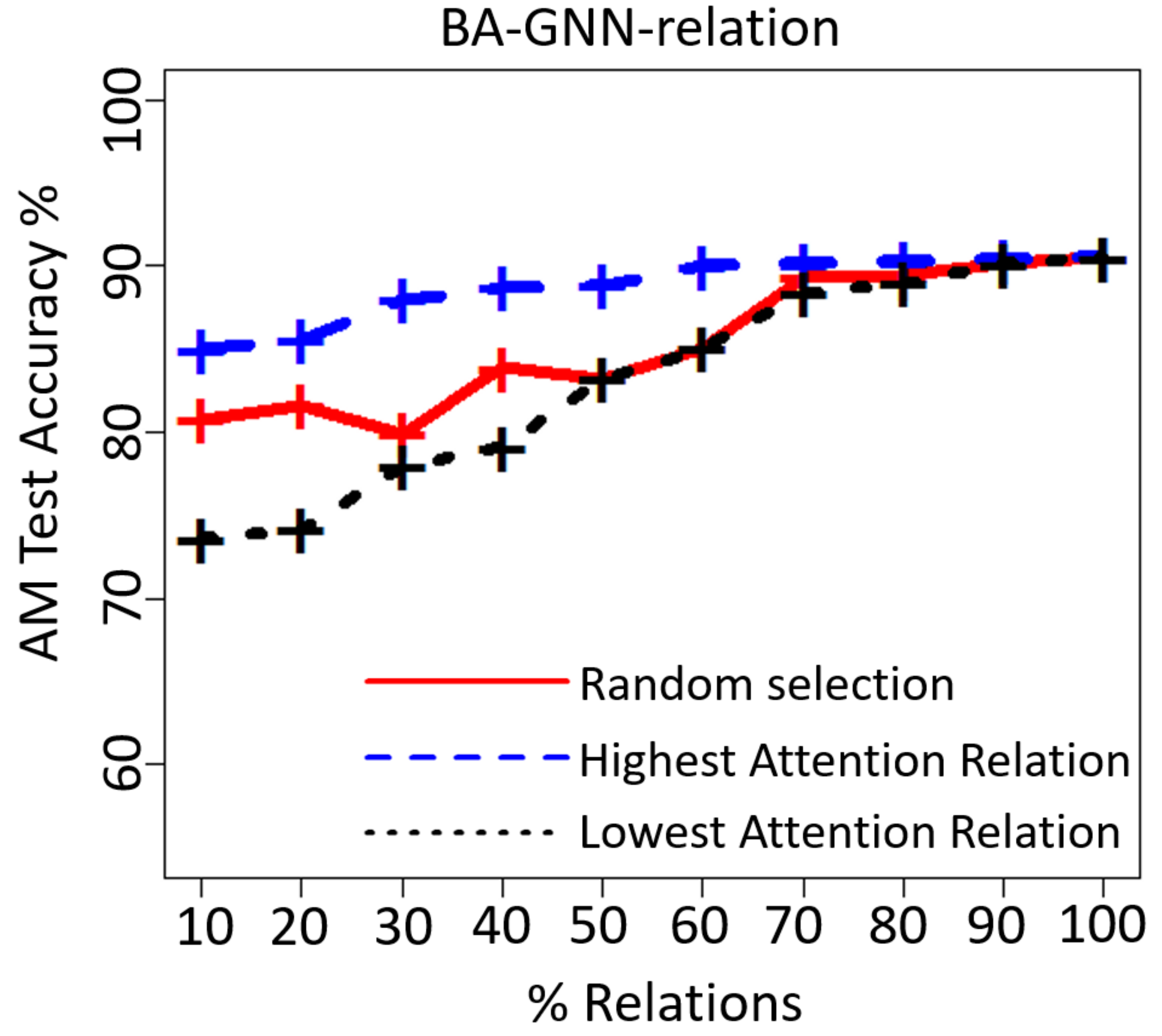}}
    {\includegraphics[width=2.2cm]{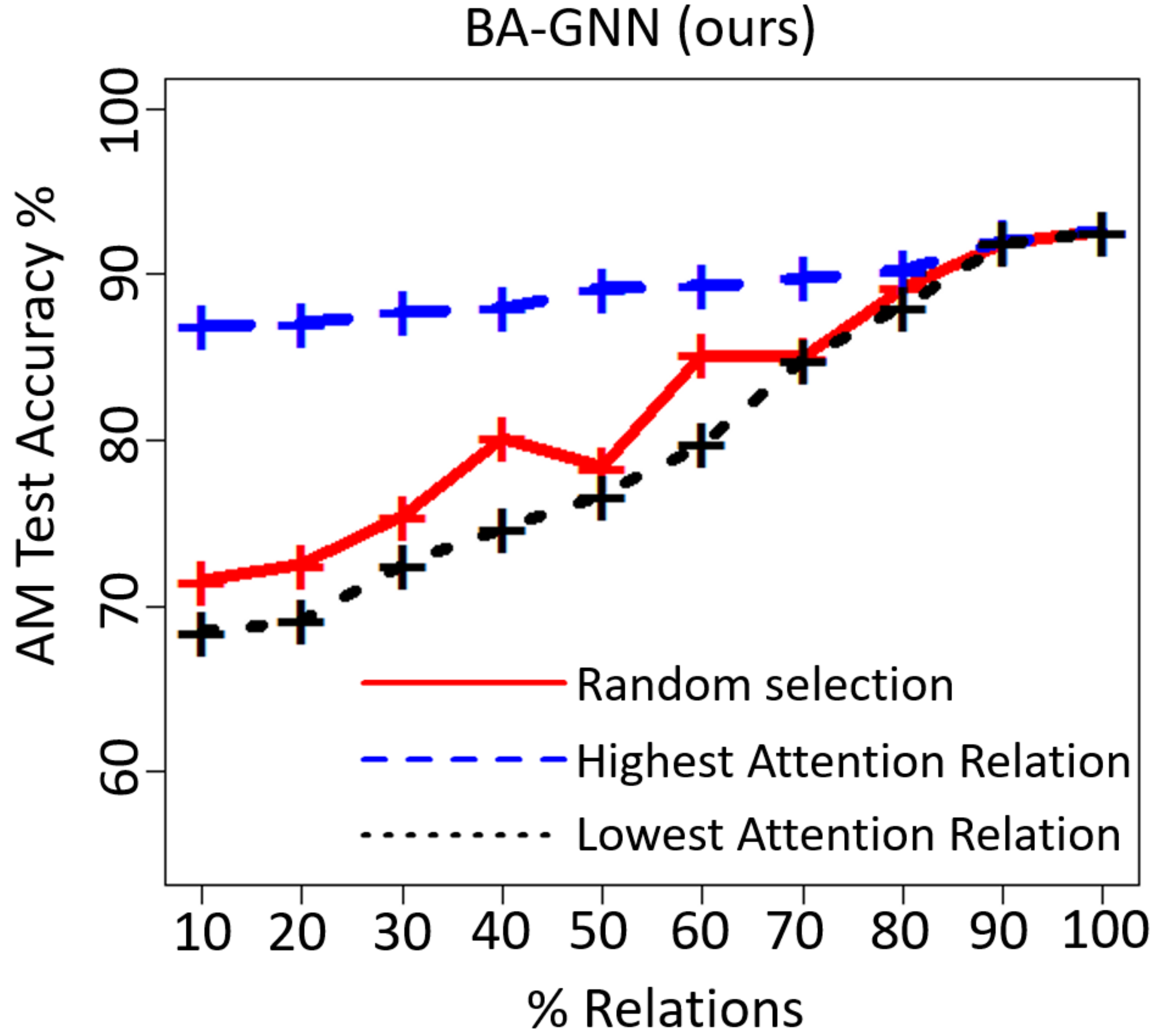}}} \vspace{-3mm} \\
    \subfloat[\centering pp1 heat map: \BAGNNSS's relation-level attention]
    {{\includegraphics[width=3.5cm]{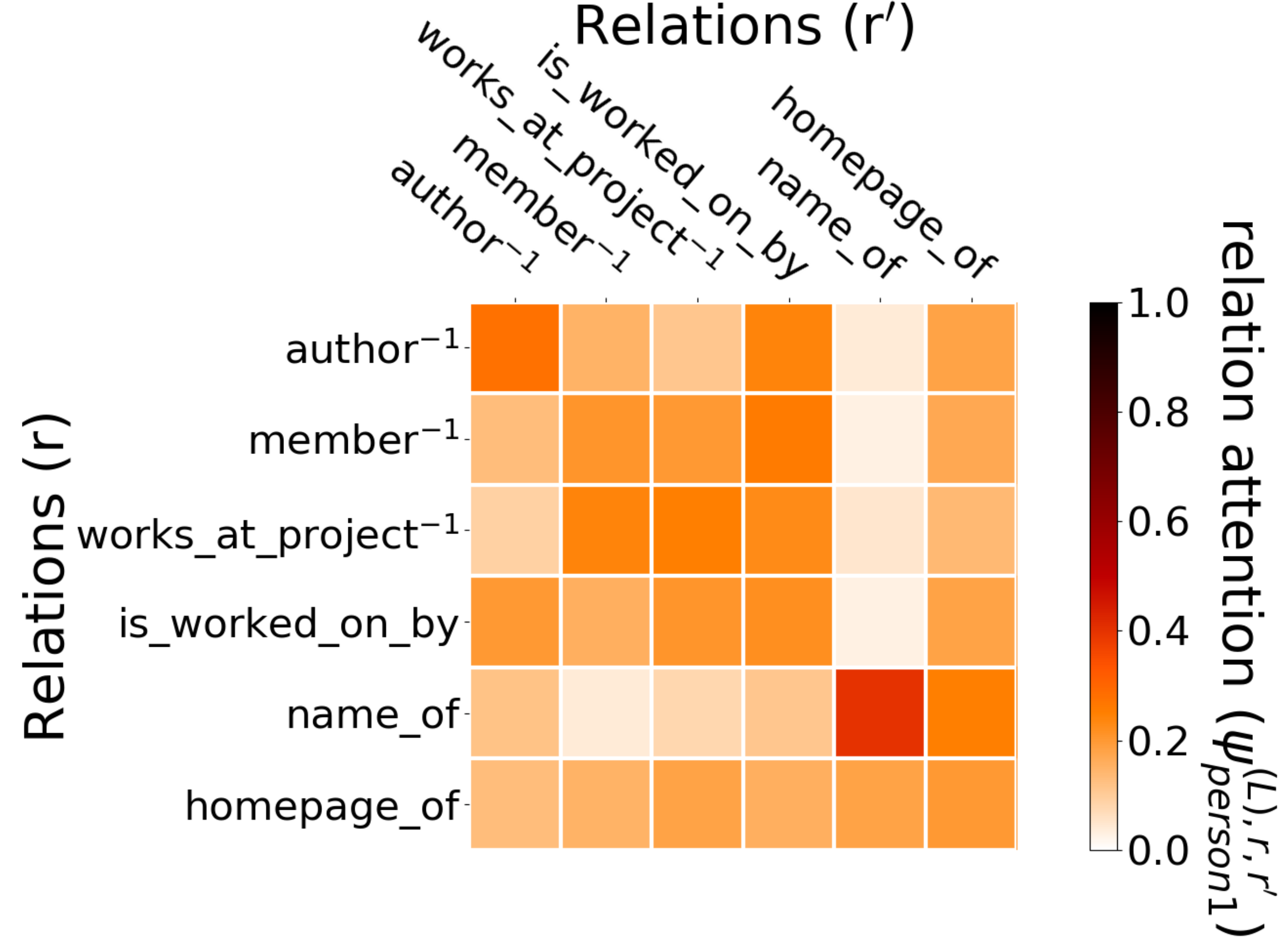} }} \hspace{2mm}%
    \subfloat[\centering pp2 heat map: \BAGNNSS's relation-level attention]
    {{\includegraphics[width=3.5cm]{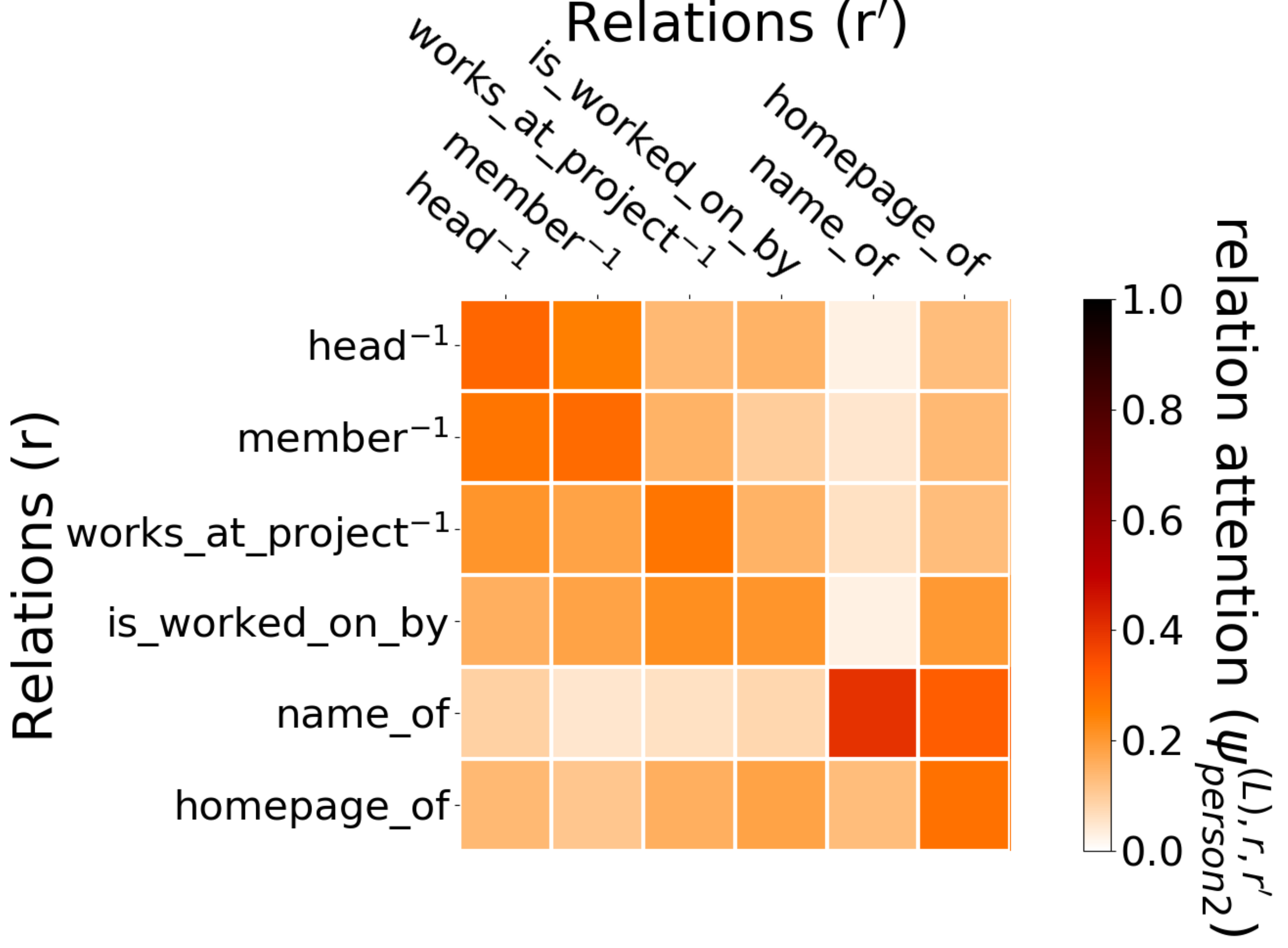} }}%
    
    \caption{\small{Experiments evaluating learned relation-level attention and graph structure of \BAGNN.}}%
    
    \label{fig:ablation-study}
    % \vspace{-7mm}
\end{figure}

\vspace{-1mm}
We conduct experiments to determine the quality of relation-level attention and graph-structure of \BAGNN. We modify the AM dataset to contain the following relation types, each with cummulative 10\% splits: (1) relations randomly selected, (2) relations with the highest relation-level attention weights from \BAGNN, and (3) relations with the lowest relation-level attention weights from \BAGNN. Experiment figures on node classification for \HAN and \BAGNN models are in Figure~\ref{fig:ablation-study}(a).

(2)'s graph structure yields the highest test accuracy on all splits of AM compared to (1) or (3), while (3) yields the lowest test accuracy. (1) is as expected in between test accuracies of (2) and (3). Models that do not learn relation-level attention (\BAGNNnode) still benefit from the graph structure identified by (2). This suggests that \BAGNN's relation-level attention can selectively identify important graph components and that its learning of graph structure can enhance other leading GNNs. 

\vspace{-3mm}

\vspace{-5mm}
\subsection{Attention Visualization}
We randomly sample two nodes belonging to entity \textit{Persons} on AIFB and plot its learned relation-level attention weights from layer $l = L$ using heat maps, seen in Figures~\ref{fig:ablation-study}(b) and (c). The corresponding partial graphs of \textit{person 1} and \textit{person 2} are in Figure~\ref{fig:aifb-example}. In Figure~\ref{fig:ablation-study}(b), \textit{author$^{-1}$} and \textit{member$^{-1}$} have high attention to \textit{is\_worked\_on\_by} because a person is likely to have publications and research affiliations in their research area. \textit{name\_of} has high attention to \textit{homepage\_of}, observed in both Figures~\ref{fig:ablation-study}(b) and (c), because a homepage may directly contain personal identifying information. In Figure~\ref{fig:ablation-study}(c), \textit{head$^{-1}$} and \textit{member$^{-1}$} have high attention to each other, since head of a research group is a member. Further, members of the research group are likely to work on the group's projects and focus on a particular research domain, explaining why \textit{works\_at\_project$^{-1}$} has higher attention to \textit{head$^{-1}$} and \textit{member$^{-1}$}, and \textit{works\_at\_project$^{-1}$} and members of the research group also have higher attention to  \textit{is\_worked\_on\_by}.

\vspace{-2mm}
\section{Conclusion}
\label{conclusions}
\vspace{-1mm}
We propose Bi-Level Attention Graph Neural Networks (\BAGNN) for modeling multi-entity and multi-relational large-scale heterogeneous graphs (HeGs), via entity type and meta relation information to learn graph structure and properties. Further, \BAGNN distinguishes nodes and relations using a novel smart-sampling bi-level attention mechanism to guide the model when aggregating features in graph neighborhoods. We conduct extensive experiments on seven real-world heterogeneous datasets, and show \BAGNN learns effective and efficient embeddings. We observe that \BAGNN outperforms all state-of-art GNN baselines on various information recovery tasks.

\vspace{-2mm}
\section{Acknowledgements}
\vspace{-1mm}
This work was supported by NSF 1705169, 1829071, 1937599, 2031187, 2106859; NIH R35-HL135772, NIBIB R01-EB027650; DARPA HR00112090027; Okawa Foundation Grant; and Amazon Research Awards. We also thank Kai-Wei Chang and Yunsheng Bai for helpful discussions.
% \vspace{-2.5mm}

\fontsize{9.0pt}{10.0pt} \selectfont
\setlength{\parskip}{0em}
\setlength{\itemsep}{0pt}
\bibliographystyle{abbrv}
\vspace{-2mm}
\small{\bibliography{main}}

\end{document}